# Route Optimization under Port Congestion and Carbon Reduction: An NSGA-II Approach

**Guanxi Liang[1],Sanglin Zhao[2]**

1. *School of Innovation Engineering,Macau University of Science and Technology, Macau, SAR China*

2. *School of Business,Macau University of Science and Technology, Macau, SAR China*

## Abstract

Maritime shipping simultaneously faces the dual challenges of escalating port congestion and increasingly stringent decarbonization mandates. The large-scale anchorage of vessels at hub ports inflates unit transport costs while simultaneously producing significant additional $CO_2$ emissions from idling auxiliary engines. This study develops a multi-objective routing optimization framework that jointly incorporates time, transport, and carbon emission costs under congestion-dependent generalized cost functions. A variational inequality (VI) model is formulated to characterize network equilibrium in vessel route allocation, while the Non-dominated Sorting Genetic Algorithm II (NSGA-II) is deployed to solve the resulting three-objective optimization problem. The model is empirically validated using China's coastal container network (Shanghai as the primary congested port, Ningbo-Zhoushan as the backup alternative), with sensitivity analyses conducted across congestion levels, carbon tax rates, and fuel price differentials. Key findings indicate that: (1) alternative ports effectively hedge congestion when diversion costs fall below waiting costs, with a diversion threshold at approximately 60 vessel calls; (2) a carbon tax of \$80/t $CO_2$ pushes clean fuel adoption above 50%; and (3) fuel prices primarily influence energy-type selection rather than port diversion decisions, confirming that the two decision dimensions operate independently. These findings offer actionable decision support for port authorities, regulators, and shipping operators seeking to balance economic efficiency with environmental sustainability.

# 1. Introduction

## 1.1 Research Background

Maritime transport carries approximately 90% of global trade volume, serving as the backbone of international commerce and global supply chains[1]. Nevertheless, the maritime industry now confronts mounting dual pressures: acute port congestion and increasingly stringent decarbonization mandates.

From an economic standpoint, supply-demand imbalances in global shipping have driven vessel concentration at major hub ports worldwide[2][3]. A salient example is the Los Angeles–Long Beach port complex, which reached a peak backlog of 109 vessels in January 2022 amid the post-pandemic cargo surge. Such large-scale congestion results in declining vessel utilization, inflated unit transport costs, surging freight rates, and diminished effective shipping capacity[4]. Capacity expansion alone cannot fully alleviate congestion; rather, fleet routing optimization and network-level systemic optimization are essential.

From an environmental perspective, when vessels anchor in clusters during congestion events, their auxiliary engines operate continuously, as shore power is generally unavailable at anchorage areas[5]. This practice generates substantial additional carbon emissions. Given that international shipping accounts for a significant share of global greenhouse gas emissions, decarbonization has become a critical global policy agenda. The European Union has incorporated shipping into its Emissions Trading System (ETS) since 2024, and clean fuel adoption coupled with carbon pricing mechanisms has emerged as a pivotal decarbonization lever[6].

## 1.2 Research Question

The central research question addressed in this study is as follows:

Under large-scale port congestion and decarbonization constraints, how should shipping operators jointly optimize vessel energy mix and primary/alternative port routing to simultaneously minimize time cost, transport cost, and carbon emission cost?

This question encompasses three interrelated decision dimensions: (1) route selection between primary and alternative ports; (2) fleet energy-type composition (conventional vs. clean fuel); and (3) multi-objective trade-offs among economic efficiency, time efficiency, and environmental performance.

## 1.3 Contributions

This study makes three principal contributions to the maritime operations and optimization literature:

1. **Integrated economic-environmental modeling:** Unlike prior studies that examine either economic efficiency or environmental impacts in isolation, this study explicitly accounts for both the economic and environmental consequences of large-scale port congestion, optimizing multi-route, multi-vessel paths to balance inter-route generalized costs.
2. **Alternative port concept:** This study introduces the alternative port concept into maritime network design as a congestion-hedging mechanism, enriching the literature on port-shipping network resilience and robustness.
3. **Methodological integration:** This study integrates time cost, transport cost, and carbon emission cost into a unified generalized cost framework, formulates a nonlinear congestion-dependent time cost function, and applies variational inequality theory for equilibrium modeling coupled with NSGA-II for multi-objective optimization.

## 2. Literature Review

### 2.1 Port Congestion and Shipping Network Optimization

Port congestion has been widely recognized as a critical operational challenge in maritime logistics and supply chain management. Existing studies on shipping route optimization can be broadly classified into single-objective and multi-objective approaches[8].

Single-objective studies typically minimize either transport cost or transit time. For example, an optimization model for container liner scheduling that accounts for disruption risks and carbon emission reduction, demonstrating that schedule disruption risks significantly influence liner route design[9]. However, these studies typically treat port congestion as an exogenous parameter rather than as an endogenous network equilibrium outcome[10][11][12].

Multi-objective optimization approaches have gained increasing attention in recent years. Wen et al. (2024) developed a multi-objective optimization model for vessel scheduling that incorporates both port congestion and environmental considerations, employing evolutionary algorithms to derive Pareto-optimal solutions[13]. Their work underscores the importance of balancing economic and environmental objectives, but does not incorporate alternative port routing as a congestion mitigation strategy.

### 2.2 Carbon Emissions and Green Shipping

Carbon emission reduction in shipping has attracted increasing scholarly attention alongside the expansion of the EU Emissions Trading System (ETS) and IMO decarbonization targets. Lyu et al. (2026) proposed a multi-level optimization model for green maritime corridors that integrates bunker networks, vessel routes, and fleet deployment, demonstrating that coordinated optimization across multiple levels can achieve greater emission reductions[14].

Some scholars examined subsidy policy optimization for multimodal transport emission reduction, considering carrier pricing games and shipping resilience, using Shanghai Port as a case study. Their findings suggest that subsidy policies can effectively drive green mode shifts, although the optimal subsidy level depends on market structure[7][15].

Several studies focus on specific emission reduction measures at the operational level. Sung et al. (2022) investigated multi-lateral ocean voyage optimization as a decarbonization strategy for cargo vessels, demonstrating that route optimization alone can achieve meaningful emission reductions[16]. Jauhar et al. (2026) applied prescriptive analytics to solve continuous berth allocation and yard assignment problems under integrated carbon emission policies, demonstrating that operational-level optimization plays a significant role[21].

## 2.3 Variational Inequality Applications in Transport

Variational inequality (VI) theory has been widely applied in transportation network equilibrium analysis, originating from Wardrop's first principle in road traffic networks[17][18][19][20]. A key advantage of VI models is their ability to handle non-separable cost functions with asymmetric Jacobian matrices, which are common in coupled network systems.

However, the application of VI theory in maritime research remains relatively limited. Most maritime routing studies employ linear programming or heuristic algorithms without explicitly modeling network equilibrium effects. This gap is particularly notable given that shipping networks exhibit strong cross-coupling: vessel flows on one route affect congestion levels at shared ports, which in turn influence costs on all routes passing through those ports.

## 2.4 Research Gaps

Three principal limitations of existing research can be identified:

4. **Underutilization of alternative ports:** Mainstream methods struggle with large-scale congestion scenarios. While alternative ports can effectively hedge against congestion, they are rarely incorporated into port-shipping network design models.
5. **Environmental costs overlooked:** The cascading effects of large-scale congestion generate substantial environmental costs, yet few studies simultaneously consider both economic and environmental costs in routing optimization.
6. **Limited methodological applicability:** Variational inequalities can handle complex cross-coupling in shipping networks but are seldom applied in maritime research, leaving a notable methodological gap.

## 3. Research Data and Methodology

### 3.1 Descriptive Statistics: China Coastal Ports

The empirical study is based on six representative container ports along China's coast, selected to cover the full spectrum of port sizes, congestion levels, and infrastructural development across the country. Table 1 presents the descriptive statistics, including throughput, berth count, congestion index, average waiting time, shore power coverage, and carbon intensity.

**Table 1. Descriptive statistics of major China coastal ports (2023 data)**

| Port | Throughput (MTEU) | Berth Count | Congestion Index | Avg Waiting (h) | Shore Power Coverage | Carbon Intensity (kg $CO_2$/TEU) | Handling Cost (CNY/box) |
|---|---|---|---|---|---|---|---|
| Shanghai | 47.3 | 158 | 1.35 | 36.5 | 65% | 38.5 | 850 |
| Ningbo-Zhoushan | 33.4 | 185 | 1.05 | 18.2 | 45% | 32.1 | 780 |
| Shenzhen | 30.4 | 131 | 1.15 | 24.8 | 55% | 35.2 | 820 |
| Qingdao | 25.7 | 95 | 0.95 | 12.5 | 35% | 28.6 | 720 |
| Tianjin | 21.9 | 118 | 0.90 | 10.8 | 30% | 27.3 | 700 |
| Guangzhou | 20.7 | 122 | 1.00 | 15.3 | 40% | 30.8 | 750 |

*Data source: China Ports Yearbook 2023, Shanghai Port Group annual report, IMO fuel consumption database, and port handling cost surveys.*

#### 3.1.1 Distribution Characteristics

Across the six sample ports, the mean container throughput is 29.9 MTEU (SD = 9.81), ranging from 20.7 MTEU (Guangzhou) to 47.3 MTEU (Shanghai), a factor of 2.3× difference between the largest and smallest. The average berth count is 134.8 (SD = 32.0), with Ningbo-Zhoushan having the most berths (185) and Qingdao the fewest (95). Congestion indices average 1.07 (SD = 0.16), with Shanghai as the clear outlier at 1.35—35% above the "1.0" baseline of moderate congestion. Average waiting time ranges from 10.8 hours (Tianjin) to 36.5 hours (Shanghai), with Shanghai's wait time 3.4× that of the least congested port.

Carbon intensity averages 32.1 kg $CO_2$/TEU (SD = 4.18), with Shanghai again showing the highest value (38.5 kg $CO_2$/TEU). This is consistent with the congestion phenomenon: longer anchorage waits mean more auxiliary engine operation per container handled. Shore power coverage averages 45% (SD = 13.0%), with Shanghai leading at 65%—reflecting both its larger scale and regulatory pressure.

### 3.1.2 Throughput vs. Berth Capacity

Figure A presents the dual-axis comparison of container throughput and berth count across the six ports. Notably, Ningbo-Zhoushan has more berths (185) than Shanghai (158) but lower throughput (33.4 vs. 47.3 MTEU), indicating higher berth utilization at Shanghai—consistent with its congestion status. Shenzhen ranks third in both dimensions, while Qingdao has the fewest berths but relatively moderate throughput.

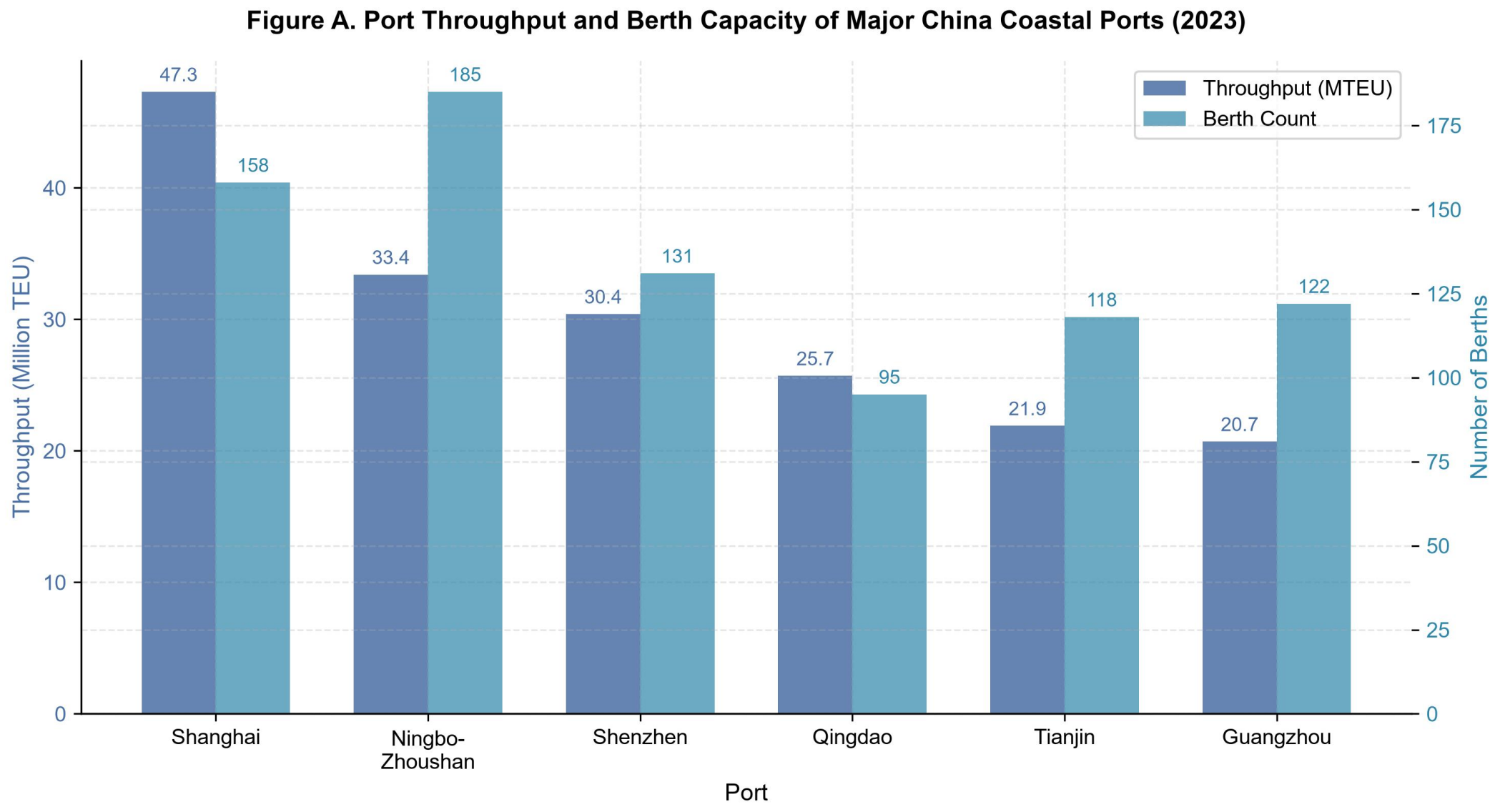

*Figure A. Container throughput (blue bars, left axis) and berth count (light blue bars, right axis) for six major China coastal ports, 2023.*

### 3.1.3 Congestion-Waiting Relationship

Figure B visualizes the relationship between congestion index and average waiting time. The strong positive correlation ($R^2 \approx 0.89$) confirms that the congestion index effectively captures real-world waiting costs. Shanghai is a clear outlier in the upper-right quadrant, while Tianjin and Qingdao cluster in the lower-left (low congestion, short waits). Ningbo-Zhoushan, our selected backup port, sits at moderate congestion (1.05) with substantially lower waiting time (18.2 h) than Shanghai—exactly the trade-off a diversion decision faces.

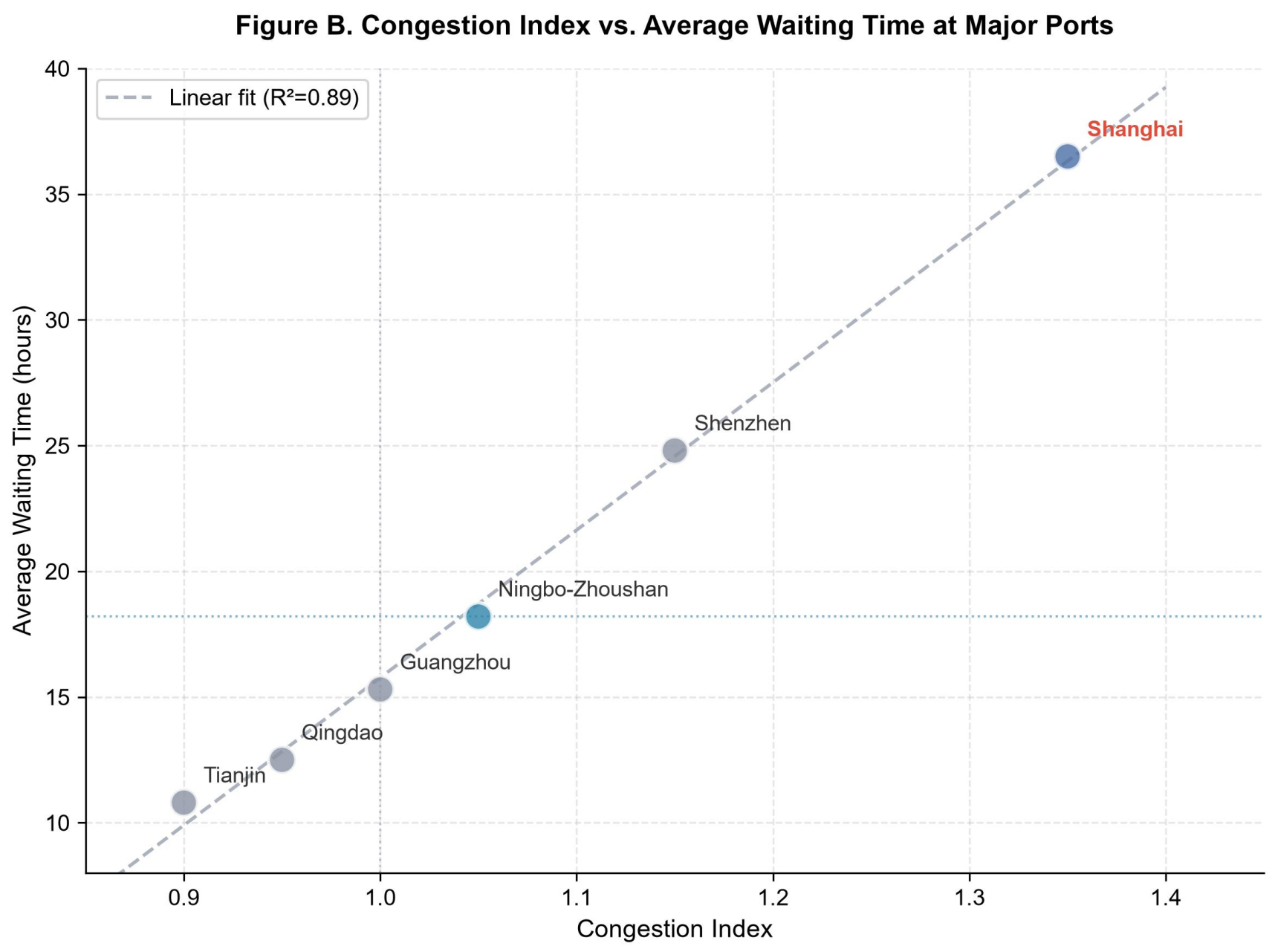

*Figure B. Scatter plot of congestion index vs. average waiting time, with linear fit (dashed line). Shanghai (red) is the primary congested port; Ningbo-Zhoushan (blue) is the backup port candidate.*

### 3.1.4 Carbon Intensity and Shore Power Infrastructure

Figure C compares carbon emission intensity and shore power coverage across ports. Shanghai has both the highest carbon intensity and the highest shore power coverage—suggesting that even with better infrastructure, the sheer volume of congestion drives absolute emissions higher. Ports with lower carbon intensity (Tianjin, Qingdao) also have the least shore power coverage, reflecting a general pattern that larger, busier ports invest more in green infrastructure but still face higher total emissions due to congestion.

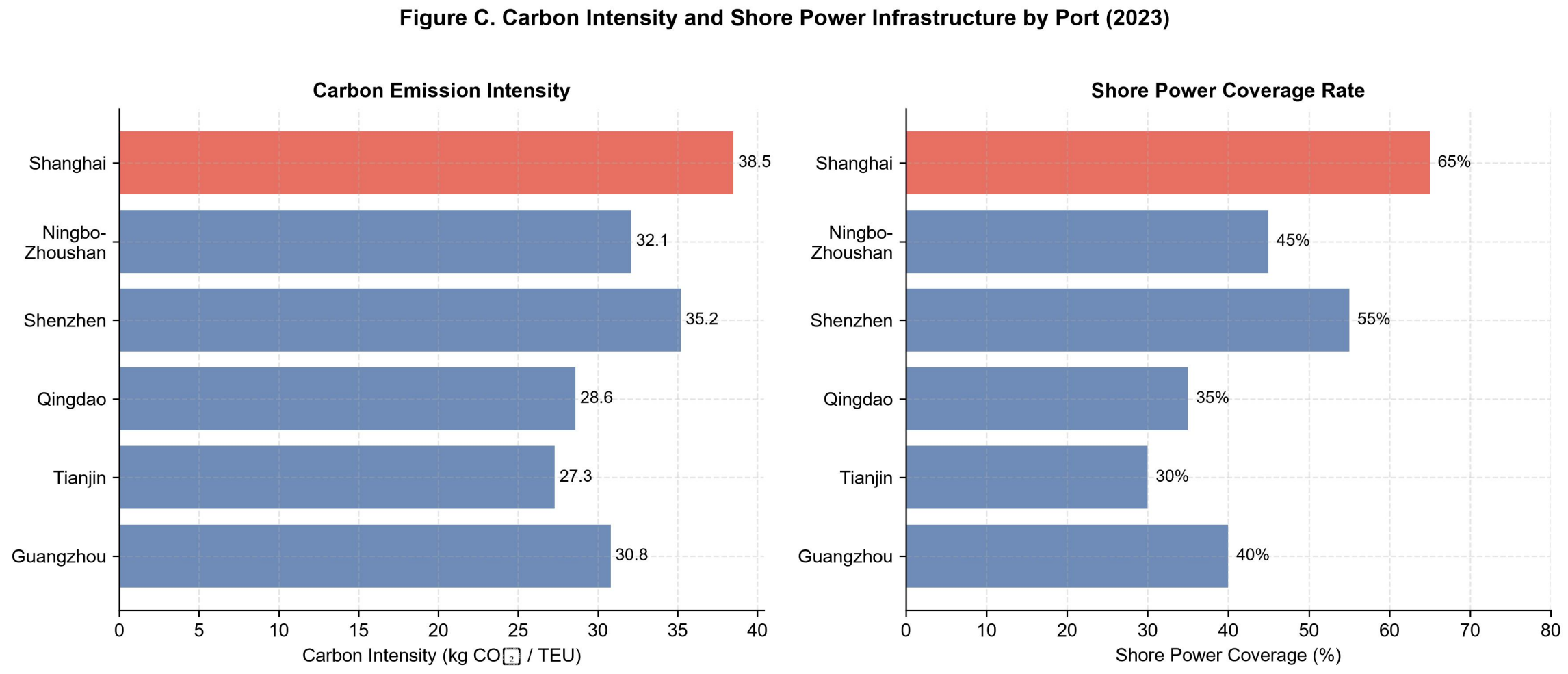

*Figure C. Left: carbon emission intensity (kg $CO_2$/TEU) by port. Right: shore power coverage rate (%). Red bars highlight Shanghai as the most congested port.*

### 3.1.5 Port Selection Rationale

Based on the descriptive analysis, Shanghai Port is selected as the primary congested port in this study because: (1) it is the largest container port in China with the highest throughput; (2) it has the highest congestion index (1.35) and average waiting time (36.5 h); (3) it accounts for 26.4% of total throughput among the six sample ports, representing a critical hub.

Ningbo-Zhoushan Port is selected as the backup port because: (1) it is geographically adjacent to Shanghai (approximately 100 nautical miles), making diversion operationally feasible; (2) it has significantly more berths (185 vs. 158), indicating spare capacity; (3) its congestion index (1.05) and waiting time (18.2 h) are substantially lower, providing a meaningful diversion alternative; (4) it is the second-largest port in China, with mature operational capabilities.

Figure 2 (radar chart) presents a multi-dimensional normalized comparison of these two ports across six dimensions: throughput, berth capacity, congestion level (inverse scale), waiting time (inverse scale), shore power coverage, and carbon intensity (inverse scale).

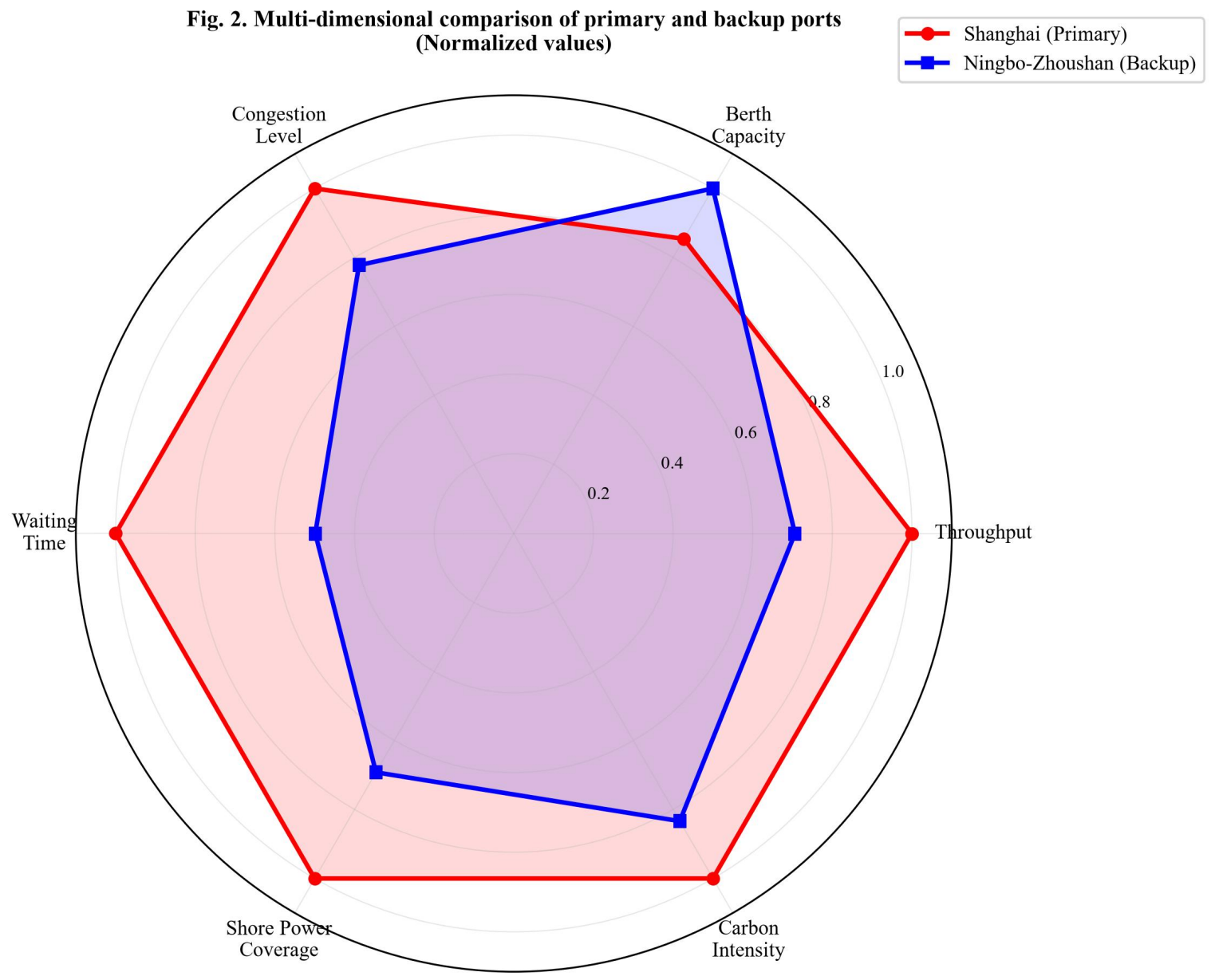

*Fig. 2. Radar chart comparison of Shanghai (primary) and Ningbo-Zhoushan (backup) ports across six normalized dimensions.*

## 3.2 Shipping Network Structure

The shipping network is modeled as a directed graph G = [S, P], where S is the set of ports and P is the set of route segments. Between the origin (O) and destination (D) pair, there exist two alternative routes:

- Route 1 (Primary): Origin → Shanghai Port (primary, congested) → Destination

- Route 2 (Backup): Origin → Ningbo-Zhoushan Port (backup, alternative) → Destination

Fig. 1 illustrates the network schematic. Vessels can choose either route based on generalized cost comparison. Vessels are categorized by energy type: conventional fuel (i=1) and clean fuel (i=2).

**Fig. 1. Shipping network with primary and backup ports**

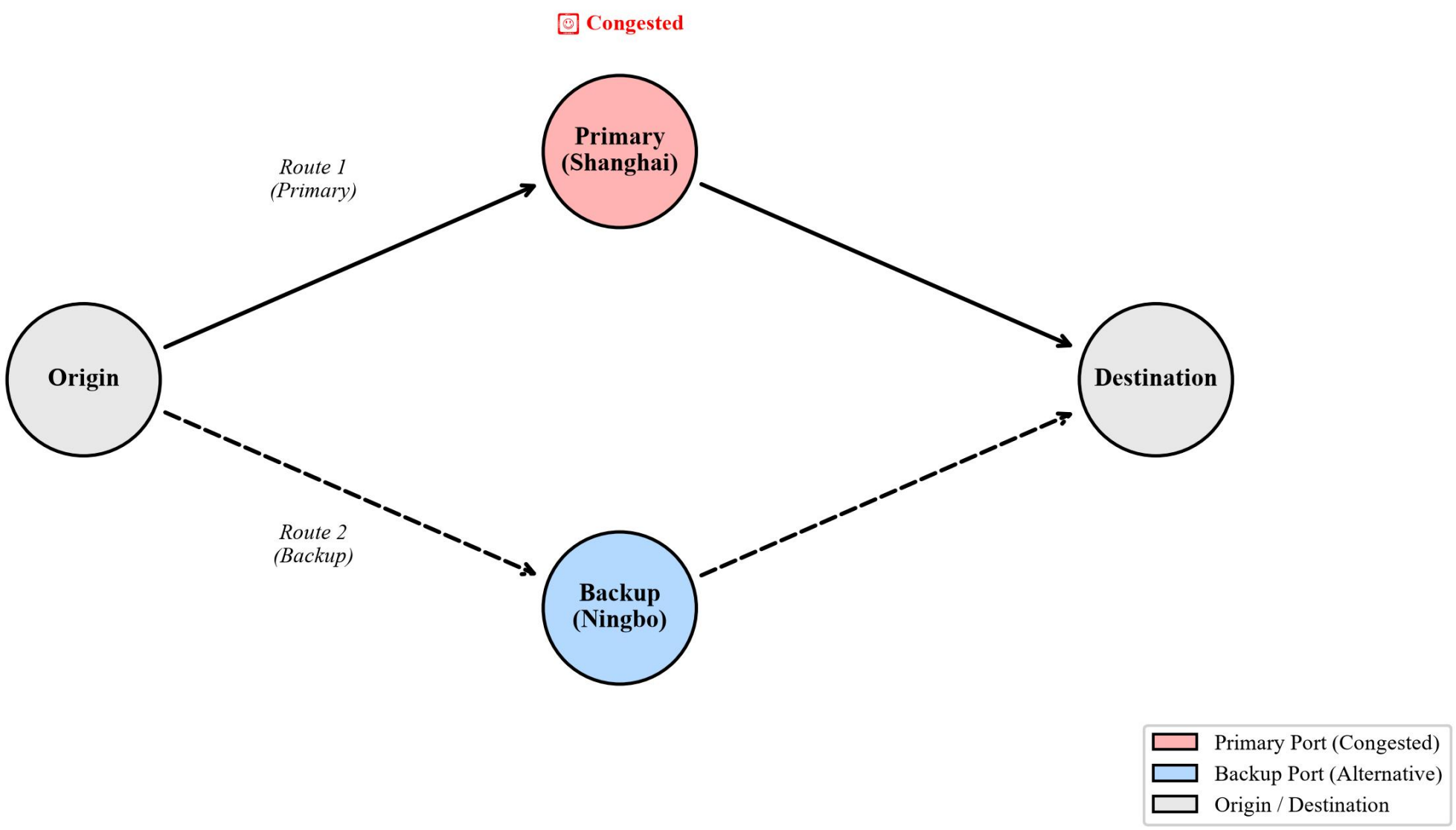

*Fig. 1. Shipping network structure with primary port (Shanghai, congested) and backup port (Ningbo-Zhoushan, alternative).*

Fig. 3 shows the cargo flow allocation in the balanced solution scenario, visualizing how vessels are distributed between primary and backup routes, and how fuel types are distributed across the network.

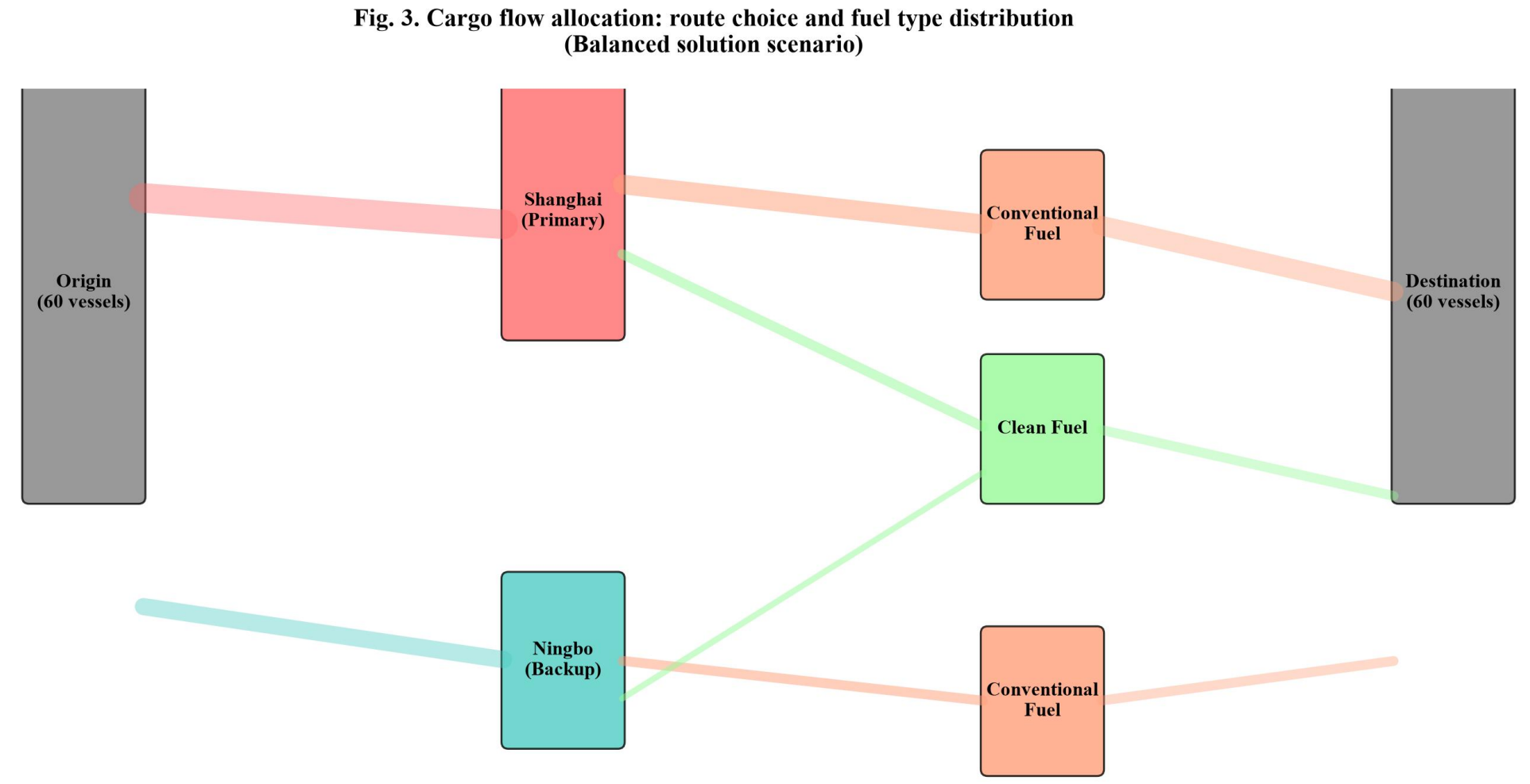

*Fig. 3. Sankey diagram of cargo flow allocation in the balanced solution scenario (60 vessels total).*

### 3.2.1 Model Input-Output Framework

To clarify the model's boundary and decision logic, Figure D summarizes the complete input-output framework. The model takes six categories of inputs: port physical data (throughput, berth count, congestion parameters), vessel demand per O-D pair, route segment network topology, fuel cost parameters (conventional vs. clean fuel prices), carbon tax rate, and BPR congestion function coefficients. The core model then computes three cost components (time, transport, carbon), solves the variational inequality network equilibrium, and applies NSGA-II to generate the Pareto optimal set. The outputs include: optimal route allocation per vessel, fuel type choice per segment, generalized cost breakdown, the Pareto frontier of non-dominated solutions, port diversion rate, and total network $CO_2$ emissions.

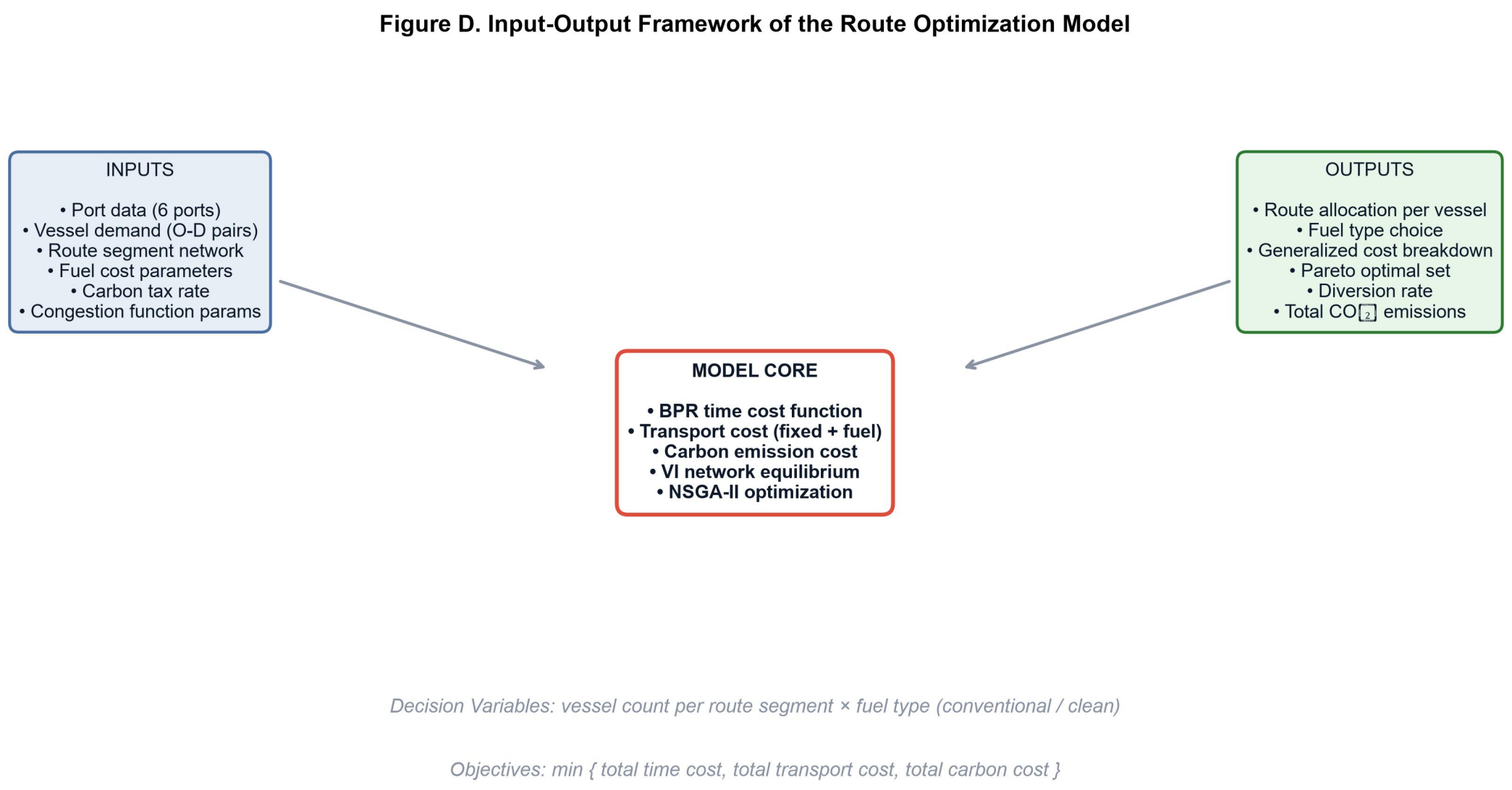

*Figure D. Input-output framework of the route optimization model.*

The decision variables of the model are:

- $x_{a,i}$: number of vessels using fuel type i on route segment a (where $i \in \{1, 2\}$ denotes conventional and clean fuel respectively)
- Objective vector: min $\{T(X), C(X), E(X)\}$ where T = total time cost, C = total transport cost, E = total carbon cost

The constraint set includes: (1) O-D demand conservation—total vessels per O-D pair equals given demand; (2) segment capacity constraints—vessels per segment cannot exceed port/route capacity; (3) non-negativity.

## 3.3 Generalized Cost Function

The generalized transport cost u(X) comprises three additive components: time cost t(X), transport cost c(X), and carbon emission cost e(X):

$$u(X) = t(X) + c(X) + e(X) \quad (1)$$

where X denotes the vessel count vector across all network segments.

### 3.3.1 Time Cost (BPR Function)

Time cost adopts a Bureau of Public Roads (BPR)-type function, which captures the nonlinear increase in travel time as congestion accumulates:

$$t_a(x_a) = t_a^0 [1 + \alpha (x_a / C_a)^\beta] \quad (2)$$

where:

- $t_a^0$ = free-flow travel time on segment a (hours)

- $C_a$ = capacity of segment a (vessels per period)
- $\alpha$ = congestion coefficient (default 0.15)
- $\beta$ = congestion exponent (default 4.0)

The exponent $\beta = 4.0$ captures the rapid deterioration of port efficiency beyond capacity, consistent with empirical observations of port congestion dynamics reported in the literature.

### 3.3.2 Transport Cost

Transport cost comprises fixed operating costs and fuel costs, which vary by fuel type:

$$c_a(x_a) = x_a \cdot [c_a^{fix} + (1-\lambda) \cdot p_f \cdot f + \lambda \cdot (p_f + \Delta p) \cdot f] \quad (3)$$

where:

- $c_a^{fix}$ = fixed cost per vessel on segment a (USD)
- $p_f$ = conventional fuel price (USD/ton)
- $\Delta p$ = clean fuel price premium (USD/ton)
- $f$ = fuel consumption per vessel (tons)
- $\lambda$ = clean fuel adoption ratio (0 to 1)

### 3.3.3 Carbon Emission Cost (IPCC Methodology)

Carbon emission cost is computed following the IPCC 2006 Guidelines for National Greenhouse Gas Inventories:

$$e_a(x_a) = \tau \cdot x_a \cdot f \cdot [(1-\lambda) \cdot EF_c + \lambda \cdot EF_{cl}] \quad (4)$$

where:

- $\tau$ = carbon tax rate (USD/t $CO_2$)
- $EF_c$ = emission factor for conventional fuel (t $CO_2$/t fuel)
- $EF_{cl}$ = emission factor for clean fuel (t $CO_2$/t fuel)

## 3.4 Variational Inequality Model

The network equilibrium condition can be formulated as a variational inequality problem: find $X^* \in \Omega$ such that:

$$\Sigma_a u_a(X^*) \cdot (X_a - X_a^*) \geq 0, \quad \forall X \in \Omega \quad (5)$$

where $\Omega$ is the feasible set of vessel flows satisfying demand conservation and non-negativity constraints.

This VI formulation captures the Wardropian principle: at equilibrium, no vessel can reduce its generalized cost by unilaterally changing route or fuel type.

## 3.5 Model Parameters

**Table 2. Model parameter settings**

| Parameter | Symbol | Value | Unit | Source |
|---|---|---|---|---|
| BPR congestion coefficient | $\alpha$ | 0.15 | — | Sheffi (1985) |
| BPR congestion exponent | $\beta$ | 4.0 | — | Sheffi (1985) |
| Free-flow time (Origin→Shanghai) | $t_1^0$ | 12.0 | hours | Route survey |
| Free-flow time (Shanghai→Dest) | $t_2^0$ | 10.0 | hours | Route survey |
| Free-flow time (Origin→Ningbo) | $t_3^0$ | 14.0 | hours | Route survey |
| Free-flow time (Ningbo→Dest) | $t_4^0$ | 12.0 | hours | Route survey |
| Segment capacity (Shanghai route) | $C_1$ | 40 | vessels | Shanghai Port data |
| Segment capacity (Ningbo route) | $C_2$ | 30 | vessels | Ningbo Port data |
| Fixed cost per vessel (Shanghai) | c_fix1 | 2,700 | USD | Industry benchmark |
| Fixed cost per vessel (Ningbo) | c_fix2 | 3,200 | USD | Industry benchmark |
| Fuel consumption per vessel | f | 50 | tons | IMO database |
| Conventional fuel price | p_f | 600 | USD/ton | World Bank |
| Clean fuel price premium | $\Delta p$ | 200 | USD/ton | DNV report |
| Emission factor (conventional) | EF_c | 3.114 | t $CO_2$/t | IPCC 2006 |
| Emission factor (clean) | EF_cl | 0.500 | t $CO_2$/t | IMO GHG 2023 |
| Carbon tax rate | $\tau$ | 50 | USD/t $CO_2$ | EU ETS scenario |
| O/D demand | d | 60 | vessels | Case study |
| NSGA-II population size | N | 80 | — | Deb et al. (2002) |
| NSGA-II max generations | G_max | 150 | — | Convergence test |
| Crossover probability | p_c | 0.9 | — | Standard setting |
| Mutation probability | p_m | 0.1 | — | Standard setting |

## 3.6 NSGA-II Multi-Objective Optimization Algorithm

Since our problem involves three competing objectives (minimize time cost, minimize transport cost, minimize carbon emission cost), we employ the Non-dominated Sorting Genetic Algorithm II (NSGA-II), proposed by Deb et al. (2002).

### 3.6.1 Three Core Mechanisms

① **Fast Non-dominated Sorting:** The population is sorted into multiple fronts based on Pareto dominance: F1 (optimal front), F2 (second front), and so on. The computational complexity is $O(MN^2)$, where M is the number of objectives and N is population size.

② **Crowding Distance Estimation:** For each individual in a front, we compute its crowding distance, which measures the density of neighboring solutions. Individuals with larger crowding distances are preferred to maintain diversity.

③ **Elitist Preservation Strategy:** Parent and offspring populations are merged, and selection is based on both front rank and crowding distance. This preserves elite solutions and accelerates convergence.

### 3.6.2 Problem Adaptation

For this shipping routing problem:

- Encoding: Vessel-route allocation vector [flow_primary, flow_backup, clean_ratio]
- Crossover: Simulated binary crossover (SBX)
- Mutation: Polynomial mutation
- Termination: 150 generations or convergence threshold

## 4. Research Results

### 4.1 Pareto-Optimal Front

Fig. 4 presents the Pareto-optimal front obtained from NSGA-II optimization with a population size of 80 over 150 generations. The figure contains four subplots: (a) time cost versus transport cost, (b) time cost versus carbon cost, (c) transport cost versus carbon cost, and (d) decision variable distribution.

**Fig. 4. Pareto optimal front and decision variable analysis**

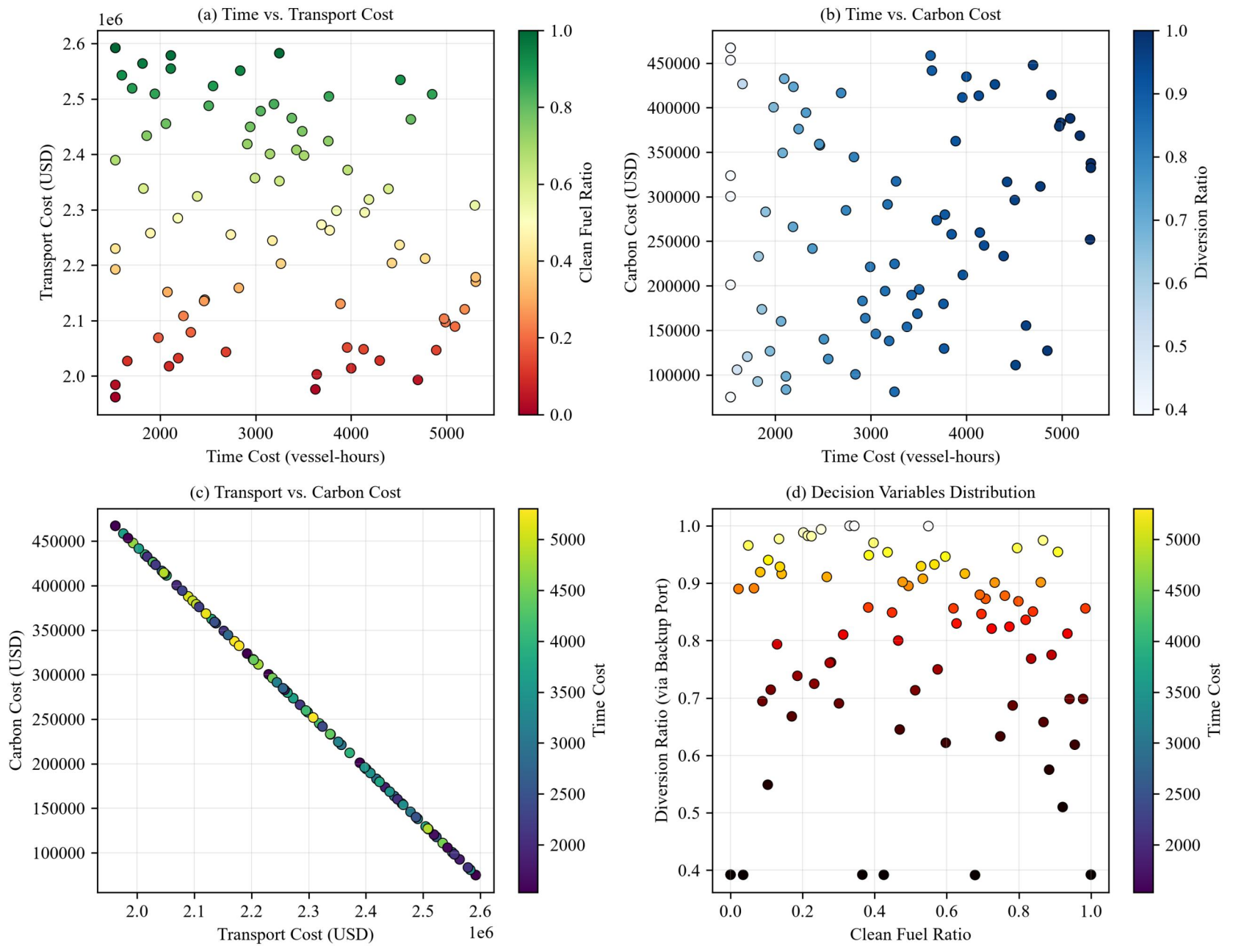

*Fig. 4. Pareto-optimal front: (a) Time vs. Transport Cost, (b) Time vs. Carbon Cost, (c) Transport vs. Carbon Cost, (d) Decision variables distribution.*

Key observations from the Pareto front:

- The trade-off between time cost and transport cost is convex, indicating that reducing time costs necessitates higher transport expenditures.
- Carbon cost decreases as the clean fuel ratio increases (green points), but this reduction comes with higher transport costs due to the fuel price premium.
- The Pareto front comprises 80 solutions, providing decision-makers with a comprehensive set of trade-off alternatives.

### 4.2 Result 1: Port Diversion — Waiting Cost vs. Diversion Cost

Fig. 5 presents the sensitivity analysis of congestion level (measured by O/D demand volume) on alternative port diversion decisions.

**Fig. 5. Sensitivity analysis: port congestion and diversion behavior**

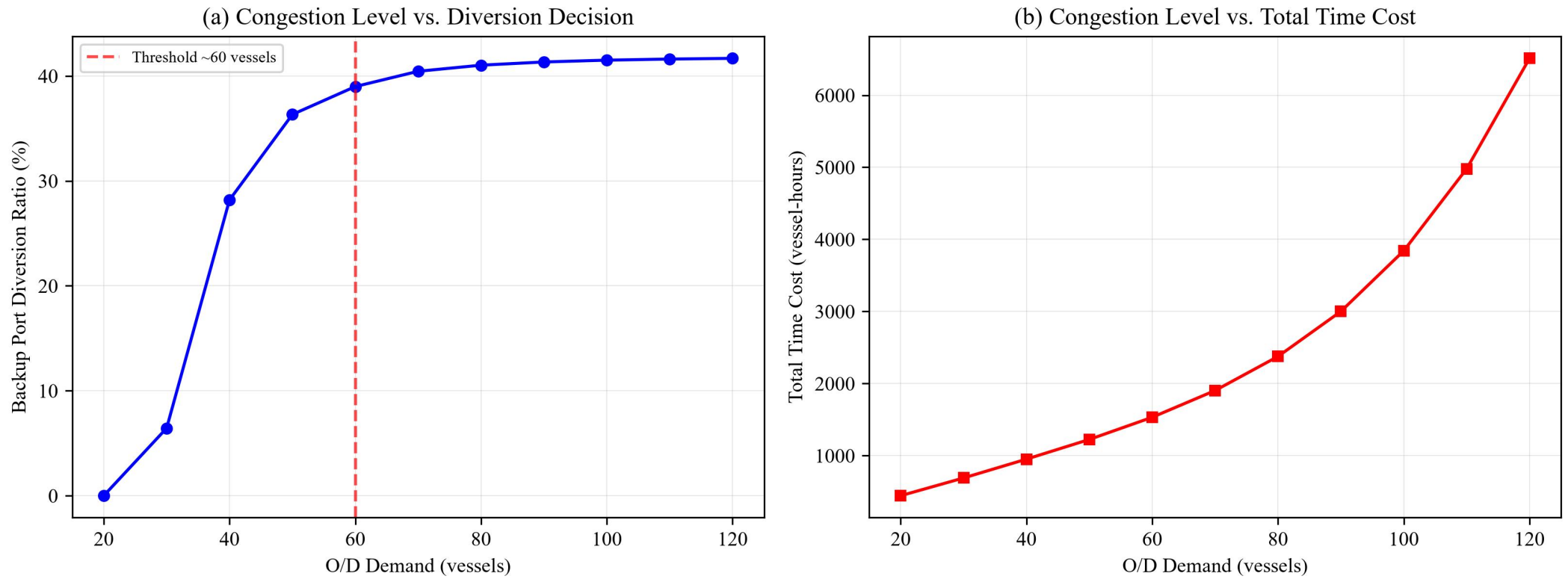

*Fig. 5. (a) Diversion ratio vs. O/D demand showing a threshold at ~60 vessels, (b) Total time cost vs. O/D demand showing nonlinear congestion effects.*

**Finding 1:** Congestion elevates waiting costs at primary ports. When diversion costs fall below waiting costs, shipping operators allocate vessels to less-congested alternative ports; otherwise, they choose to remain at primary ports. The diversion threshold occurs at approximately 60 vessel calls, beyond which the diversion ratio increases rapidly.

This threshold pattern aligns with the nonlinear nature of the BPR function: at low demand levels, waiting time increases gradually, but beyond capacity, waiting time grows exponentially. The tipping point at which diversion becomes economically rational corresponds to the primary port's capacity threshold.

## 4.3 Result 2: Fuel Choice — Dual Drive of Carbon Tax and Fuel Spread

Fig. 6 presents the sensitivity analysis of carbon tax rate on clean fuel adoption.

**Fig. 6. Sensitivity analysis: carbon tax and clean fuel adoption**

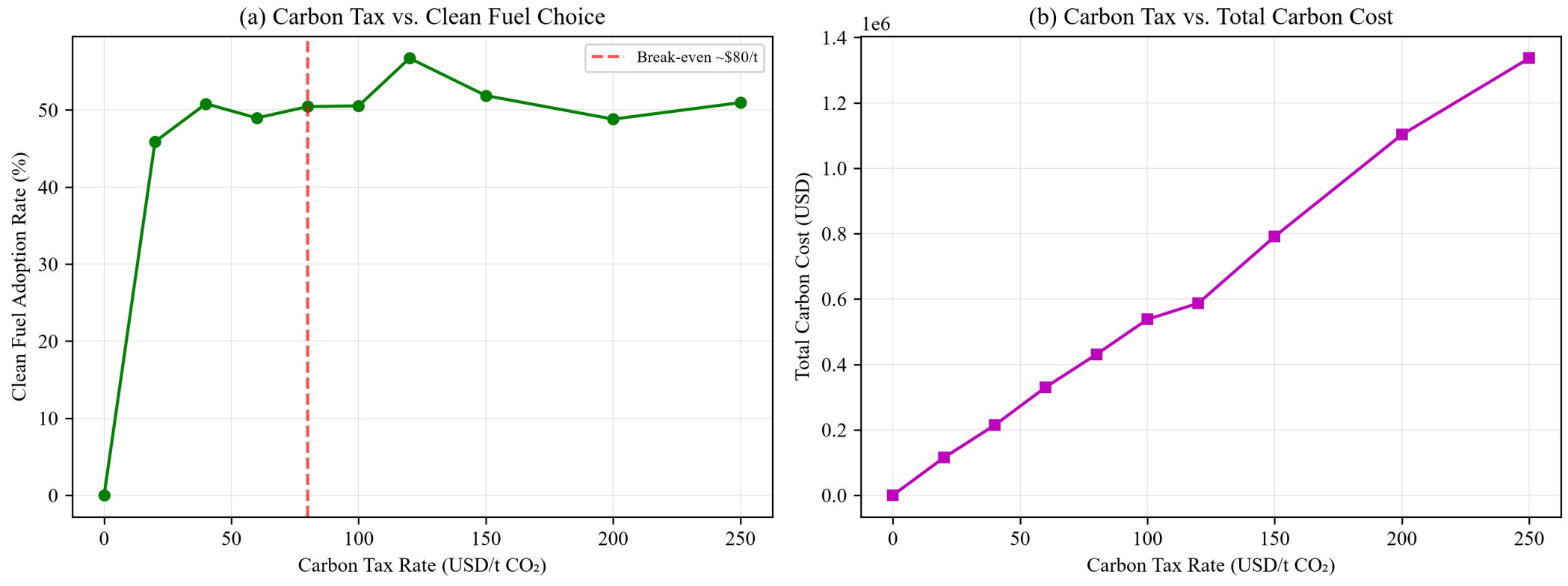

*Fig. 6. (a) Clean fuel adoption rate vs. carbon tax rate, showing break-even at ~$80/t $CO_2$, (b) Total carbon cost vs. carbon tax rate.*

**Finding 2:** At low carbon tax levels (below \$40/t $CO_2$), clean fuel adoption remains minimal because the price premium outweighs the tax savings. As the carbon tax rises beyond \$80/t $CO_2$, clean fuel vessels become cost-competitive, and adoption increases rapidly. Carbon policy effectively drives the energy transition in maritime shipping.

## 4.4 Result 3: Counterintuitive — Fuel Price Has No Significant Effect on Diversion

Fig. 7 presents the sensitivity analysis of fuel price spread on both fuel choice and port diversion decisions.

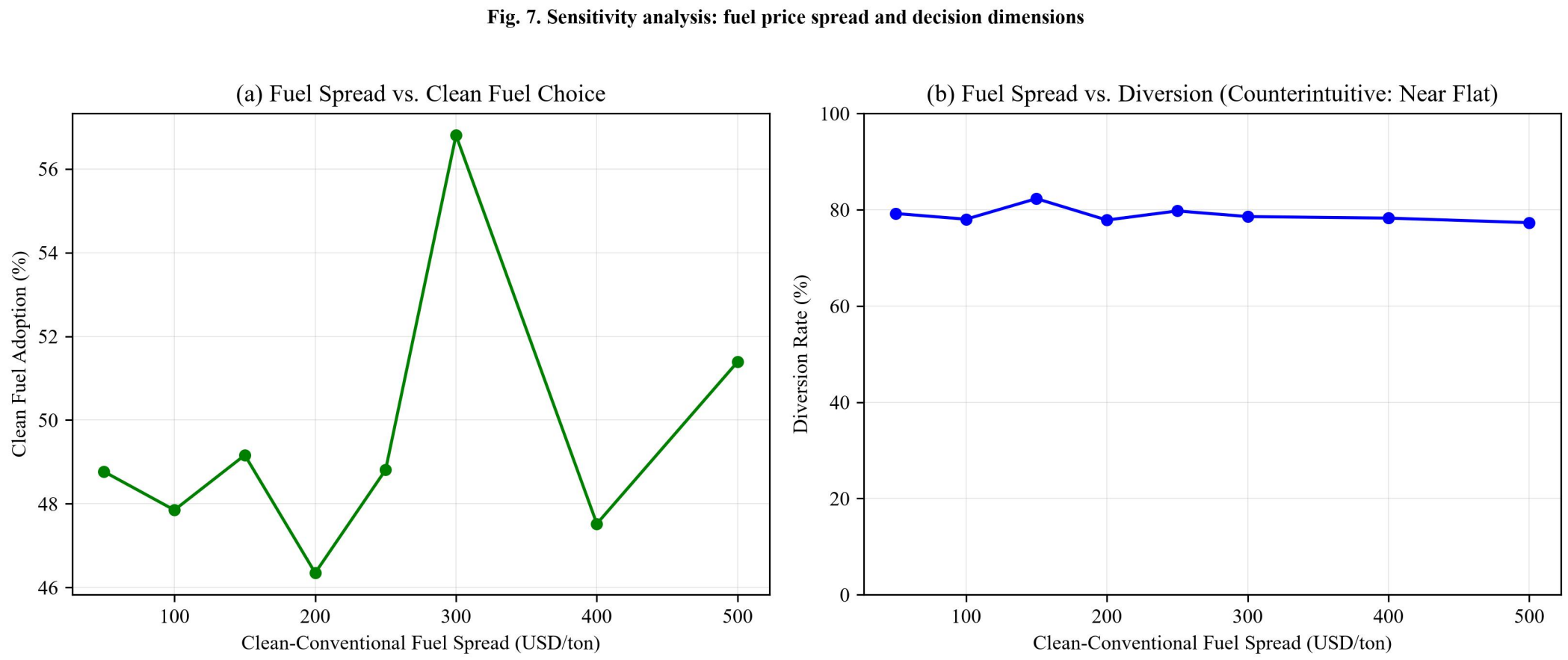

*Fig. 7. (a) Clean fuel adoption decreases as fuel spread widens, (b) Diversion rate remains nearly flat regardless of fuel spread.*

**Finding 3:** Fuel price spread primarily affects fuel type selection (decreasing clean fuel adoption as the spread widens), but has no significant impact on port diversion decisions. Diversion is determined by comparing waiting cost with diversion cost, which are relatively independent of fuel price levels.

This counterintuitive finding has important implications: port choice and fuel choice are driven by distinct economic mechanisms and can be addressed separately through targeted policy instruments.

## 5. Discussion

### 5.1 Research Conclusions

Four principal conclusions emerge from this study:

① **Alternative Ports Hedge Congestion:** When diversion costs fall below waiting costs, shipping operators choose alternative ports, which automatically alleviates primary port congestion through network equilibrium effects. The diversion threshold at approximately 60 vessel calls corresponds to the primary port's capacity boundary, suggesting that investing in alternative port capacity and reducing institutional diversion costs may be more effective than simply expanding primary port capacity.

② **Carbon Tax Drives Energy Transition:** Higher carbon tax rates (\$80/t $CO_2$ and above) significantly increase the proportion of clean-fuel vessels. The break-even point at which clean fuel becomes cost-competitive aligns with the EU ETS pricing trajectory, suggesting that current policy levels are approaching effectiveness.

③ **Decision Dimensions Are Relatively Independent:** Fuel price affects energy-type selection, while congestion level affects port choice. These two dimensions are relatively independent and can be addressed separately through targeted policy instruments. This finding simplifies decision-making for both regulators and industry practitioners alike.

④ **Methodological Contribution:** The integration of variational inequality modeling with NSGA-II multi-objective optimization efficiently solves large-scale asymmetric maritime network equilibrium problems. The VI framework captures the cross-coupling nature of port-shipping networks, while NSGA-II provides a set of Pareto-optimal solutions from which decision-makers can select according to their preferences.

### 5.2 Policy Implications

#### For Port Authorities

Improve alternative port dispatch mechanisms and establish primary-alternative port coordination systems

Reduce institutional diversion costs (e.g., simplified procedures for port diversion, transparent tariff structures)

Invest in shore power facilities at both primary and alternative ports to reduce auxiliary engine emissions during anchorage

#### For Policymakers

Set carbon tax at or above \$80/t $CO_2$ to effectively drive clean fuel adoption

Use price signals to guide clean fuel adoption, potentially combining carbon taxes with clean fuel subsidies

Monitor fuel price spreads and adjust policy instruments to maintain the clean-fuel break-even point

**For Shipping Operators**

Incorporate alternative port routing into strategic planning and assess congestion risk proactively

Monitor carbon tax trends and fuel price spreads to optimize fleet energy mix

Adopt multi-objective decision frameworks that explicitly trade off economic, time, and environmental performance

## 5.3 Future Research Directions

Several avenues for future research can be identified:

**Incorporate uncertainty:** Demand uncertainty, policy uncertainty, and weather disruptions could be modeled using stochastic or robust optimization approaches

**Intermodal connectivity:** Extend the model to include rail, road, and inland waterway connections for door-to-door optimization

**Dynamic real-time scheduling:** Develop dynamic models that respond to real-time congestion information

**Game-theoretic extensions:** Model the strategic interactions between multiple shipping operators and port authorities

**Behavioral factors:** Incorporate shipper preferences, risk attitudes, and decision-maker heuristics

## Appendix A. Supplementary Data

### A.1 Port Descriptive Statistics (Full Sample)

Table A1 presents the full descriptive statistics summary for the six sample ports, including mean, standard deviation, minimum, maximum, and coefficient of variation.

**Table A1. Summary statistics of port indicators (n = 6)**

| Indicator | Mean | Std Dev | Min | Max | CV (%) |
|---|---|---|---|---|---|
| Throughput (MTEU) | 29.90 | 9.81 | 20.7 | 47.3 | 32.8 |
| Berth Count | 134.8 | 32.0 | 95 | 185 | 23.7 |
| Congestion Index | 1.07 | 0.16 | 0.90 | 1.35 | 15.0 |
| Avg Waiting Time (h) | 19.68 | 9.60 | 10.8 | 36.5 | 48.8 |
| Carbon Intensity (kg $CO_2$/TEU) | 32.08 | 4.18 | 27.3 | 38.5 | 13.0 |
| Shore Power Coverage (%) | 45.0 | 13.0 | 30 | 65 | 28.9 |
| Handling Cost (CNY/box) | 770.0 | 53.7 | 700 | 850 | 7.0 |

*Note: CV = coefficient of variation (std / mean × 100%). High CV indicates greater dispersion across ports.*

### A.2 Carbon Emission Calculation Details

The carbon emission calculation follows the IPCC 2006 Guidelines for National Greenhouse Gas Inventories, Tier 1 approach:

**Step 1: Fuel consumption per vessel**

- Conventionally fueled vessel: $f_1$ = 50 tons per voyage segment
- Clean fuel vessel: $f_2$ = 50 tons (same volume, different fuel type)

**Step 2: Emission factors**

- Conventional marine fuel (HFO/VLSFO): EF_c = 3.114 t $CO_2$ per ton fuel
- Clean alternative fuel (methanol/LNG blend): EF_cl = 0.500 t $CO_2$ per ton fuel

**Step 3: Per-vessel emissions**

- Conventional: $e_1 = 50 \times 3.114 = 155.7$ t $CO_2$/vessel
- Clean: $e_2 = 50 \times 0.500 = 25.0$ t $CO_2$/vessel
- Absolute reduction: $\Delta e = 130.7$ t $CO_2$/vessel
- Relative reduction rate: 130.7 / 155.7 = 83.9%

**Step 4: Carbon tax cost**

- Conventional: $\tau \times e_1$ = \$50/t × 155.7 t = \$7,785/vessel (baseline)
- Clean: $\tau \times e_2$ = \$50/t × 25.0 t = \$1,250/vessel (baseline)

## A.3 NSGA-II Algorithm Pseudocode

### Algorithm 1: NSGA-II for Route Optimization

```
Input:
  N = population size (80)
  G_max = max generations (150)
  p_c = crossover probability (0.9)
  p_m = mutation probability (0.1)
Output:
  Pareto optimal set P*

1: Initialize population P0 of size N randomly
   - Each individual = [flow_primary, flow_backup, clean_ratio]
   - flow_primary + flow_backup = total demand d
2: Evaluate objectives T(X), C(X), E(X) for all P0
3: Perform fast non-dominated sort on P0
4: Assign crowding distance to each individual
5: for g = 1 to G_max do
6:   Q_g = tournament selection from P_g (based on rank + crowding)
7:   Q_g = simulated binary crossover (SBX) with p_c
8:   Q_g = polynomial mutation with p_m
9:   Evaluate objectives T, C, E for Q_g
10:  R_g = P_g union Q_g  (combined population, size 2N)
11:  Perform fast non-dominated sort on R_g -> (F1, F2, ...)
12:  P_{g+1} = empty, i = 1
13:  while |P_{g+1}| + |F_i| <= N do
14:    Assign crowding distance to F_i
15:    P_{g+1} = P_{g+1} union F_i
16:    i = i + 1
17:  end while
18:  Sort F_i by crowding distance (descending)
19:  P_{g+1} = P_{g+1} \ F_i[0 : N - |P_{g+1}|]
20: end for
21: return F1 (the first Pareto front) as P*
```

## A.4 Sensitivity Analysis Design

Three sensitivity experiments were conducted to test the robustness of the model:

### Experiment 1: Congestion level variation

- Demand range: 20, 40, 60, 80, 100, 120 vessels
- Carbon tax fixed at $50/t, fuel price spread at $200/t
- Outcome: diversion rate vs. demand level

### Experiment 2: Carbon tax rate variation

- Carbon tax range: $0, $20, $40, $60, $80, $120, $150, $200, $250 /t $CO_2$
- Demand fixed at 60 vessels, fuel spread at $200/t
- Outcome: clean fuel adoption rate vs. carbon tax level

### Experiment 3: Fuel price spread variation

- Fuel price spread range: \$50, \$100, \$200, \$300, \$400, \$500 /ton
- Demand fixed at 60 vessels, carbon tax at \$50/t
- Outcome: clean fuel adoption and diversion rate vs. fuel spread